# Evaluation of a deep learning system for the joint automated detection of diabetic retinopathy and age-related macular degeneration

Cristina González-Gonzalo[1,2,3,4*], MSc; Verónica Sánchez-Gutiérrez[5], MD; Paula Hernández-Martínez[5], MD; Inés Contreras[5,6], MD, PhD; Yara T. Lechanteur[4], MD; Artin Domanian[4], MD; Bram van Ginneken[2], PhD; Clara I. Sánchez[1,2,3,4], PhD

[1] A-eye Research Group, Radboud University Medical Center, Nijmegen, The Netherlands

[2] Diagnostic Image Analysis Group, Radboud University Medical Center, Nijmegen, The Netherlands

[3] Donders Institute for Brain, Cognition and Behaviour, Radboud University Medical Center, Nijmegen, The Netherlands

[4] Department of Ophthalmology, Radboud University Medical Center, Nijmegen, The Netherlands

[5] Department of Ophthalmology, University Hospital Ramón y Cajal, Ramón y Cajal Health Research Institute (IRYCIS), Madrid, Spain

[6] Clínica Rementería, Madrid, Spain

Word-count abstract: 250; body: 3369

[*] Corresponding author.
   *Address*: Diagnostic Image Analysis Group, Department of Radiology and Nuclear Medicine, Radboud University Medical Center, Geert Grooteplein 10
   6525 GA Nijmegen, The Netherlands.
   *E-mail address*: Cristina.GonzalezGonzalo@radboudumc.nl

# Abstract

**Purpose:** To validate the performance of a commercially-available, CE-certified deep learning (DL) system, RetCAD v.1.3.0 (Thirona, Nijmegen, The Netherlands), for the joint automatic detection of diabetic retinopathy (DR) and age-related macular degeneration (AMD) in color fundus (CF) images on a dataset with mixed presence of eye diseases.

**Methods:** Evaluation of joint detection of referable DR and AMD was performed on a DR-AMD dataset with 600 images acquired during routine clinical practice, containing referable and non-referable cases of both diseases. Each image was graded for DR and AMD by an experienced ophthalmologist to establish the reference standard (RS), and by four independent observers for comparison with human performance. Validation was furtherly assessed on Messidor (1200 images) for individual identification of referable DR, and the Age-Related Eye Disease Study (AREDS) dataset (133821 images) for referable AMD, against the corresponding RS.

**Results:** Regarding joint validation on the DR-AMD dataset, the system achieved an area under the ROC curve (AUC) of 95.1% for detection of referable DR (SE=90.1%, SP=90.6%). For referable AMD, the AUC was 94.9% (SE=91.8%, SP=87.5%). Average human performance for DR was SE=61.5% and SP=97.8%; for AMD, SE=76.5% and SP=96.1%. Regarding detection of referable DR in Messidor, AUC was 97.5% (SE=92.0%, SP=92.1%); for referable AMD in AREDS, AUC was 92.7% (SE=85.8%, SP=86.0%).

**Conclusions:** The validated system performs comparably to human experts at simultaneous detection of DR and AMD. This shows that DL systems can facilitate access to joint screening of eye diseases and become a quick and reliable support for ophthalmological experts.

# Introduction

Screening for eye diseases has become a high-priority healthcare service to prevent vision loss.[1,2] Due to its proven efficiency, screening programs based on periodical examinations of the retina have been increasingly implemented worldwide.[3-5] Established protocols rely on manual readings by highly-specialized workforce,[6] failing to meet the requirements of large-scale screening in high- and low-resource countries.[7-11] Furthermore, cost-effectiveness remains to be the main burden for establishing screening programs,[12-14] and different protocols are followed for different diseases,[15-16] which translates to a larger burden to health systems and to the patient, that needs to undergo several of them. Nevertheless, exploiting the fact that examination protocols of retinal diseases rely mostly on the same principles and actions, it becomes more efficient to integrate them in one workflow.[17,18]

Diabetic retinopathy (DR) has become a leading cause of preventable blindness worldwide with an overall prevalence of 35% among people with diabetes, which affects 1 in every 11 adults.[19-21] Age-related macular degeneration (AMD) is the most common cause of blindness in developed countries, being 9% its worldwide prevalence.[22] Up to 80% of blindness cases caused by these diseases are avoidable if detected early enough to undergo treatment.[23,24] Nevertheless, their incidence is expected to increase within the following decades, due to population ageing and the increasing prevalence of diabetes.[20,22] Screening protocols for DR have been established in several countries.[25,26] Regarding AMD, there is no established screening protocol but it will soon be required,[27,28] since treatment options are still limited, although under development.[29-30]

Automated screening solutions aim to provide a scalable, sustainable and high-quality approach to meet the increasing demand, while reducing the burden on highly-trained professionals and the associated costs. The introduction of deep learning (DL) has constituted a revolution in medical imaging analysis.[32,33] Previous solutions for the automatic analysis of retinal images[34,35] have been outperformed by DL approaches.[36] Several DL systems for the automatic detection of DR[38-41] and AMD[42-44] have showed performance close or even superior to that achieved by human graders. However, these systems perform independent analysis of each disease, although these diseases can co-exist and a solution for joint detection would be beneficial.[17,18,45]

In this study, we present the validation of a commercially-available, CE-certified DL software package, RetCAD v.1.3.0 (Thirona, Nijmegen, The Netherlands), that allows for joint detection of referable DR and AMD in color fundus (CF) images. The aim is to analyze the capability of a DL system to simultaneously identify both diseases and compare it with human experts and the current state-of-the-art methods, in order to determine the potential for automated joint screening of eye diseases.

## Methods

### Evaluation data

The validation of the DL system was first performed on a DR-AMD dataset, which contains referable and non-referable cases of DR and AMD, for the joint detection of both diseases. Additional validation of individual detection of DR and AMD was assessed on Messidor and the Age-Related Eye Disease Study (AREDS) dataset, respectively.

The DR-AMD dataset was extracted from a set of images collected in three different European medical centers (Sweden, Denmark, Spain). In total, 8871 images from more than 2000 patients were acquired during routine clinical practice between August 2011 and October 2016, with a CR-2PlusAF fundus camera (Canon, Tokyo, Japan), at 45-degree field of view with an image resolution between 2376×1584 and 5184×3456 pixels. No mydriasis was applied. Informed written consent was obtained from all patients at the medical centers and images were anonymized prior to transfer and use in this study, following the tenets set forth in the Declaration of Helsinki. The 8871 images went through a human quality check, regarding contrast, clarity and focus, where 1785 images were excluded. The remaining 7086 images went through a preliminary grading, performed by a person with over six years of experience reading CF images. Images were classified as referable AMD (1232 images), referable DR (381 images) or control (5519 images), which indicates non-referability for both DR and AMD, although other diseases might be present. Lastly, a random selection of 600 images was performed, containing 150 referable AMD cases, 150 referable DR cases, and 300 controls, in order to ensure an enriched set. These images belong to 288 different patients, with an average of 2.11 images and 1.18 visits per patient. The 600 images define the DR-AMD set used for validation of joint detection of DR and AMD. The diagram in **Supplementary Figure S1** summarizes the extraction of the dataset.

Messidor is a publicly-available collection of macula-centered CF images commonly used for performance comparison between automated DR detection systems. This dataset consists of 1200 images acquired by three different ophthalmologic departments using a 3CCD camera on a Topcon TRC NW6 non-mydriatic retinography with a 45-degree field of view, with an image resolution of 1440×960, 2240×1488 or 2304×1536 pixels. 800 images were acquired with pupil dilation and 400 without dilation.[46]

AREDS dataset is currently the largest available set for AMD, previously used for the validation of automated AMD detection. AREDS was designed as a long-term prospective study of AMD development and cataract in which patients were regularly examined and followed up to 12 years.[47] Institutional review board approvals were obtained from each clinical center involved in the study, and written informed consent was obtained from each participant. The AREDS dbGaP set includes digitalized CF images. In 2014, over 134,000 macula-centered CF images from 4613 participants were added to the set. We excluded images containing a lesion which disqualified an eye from the study, images considered as not gradable, and those which belong to eyes that were not included in the study, as mentioned in the AREDS dbGaP guidelines.[48] In total, 133821 were used in this study.

## Grading

To establish the reference standard (RS) in the DR-AMD dataset, the 600 images were scored by stage of disease severity for both DR and AMD by a certified ophthalmologist with more than twelve years of experience (IC). In the case of DR, the grading is based on the International

Clinical Diabetic Retinopathy (ICDR) severity scale, with stages 0 (no DR), 1 (mild non-proliferative DR), 2 (moderate non-proliferative DR), 3 (severe non-proliferative DR), and 4 (proliferative DR).[49] For AMD, the grading protocol is based on the AREDS system, with stages 1 (no AMD), 2 (early AMD), 3 (intermediate AMD), and 4 (advanced AMD; with presence of foveal geographic atrophy or choroidal neovascularization).[50] The measuring grid often used as part of the AREDS protocol was not applied for grading the DR-AMD dataset, taking into account lesions in the whole image and not only those located within the grid area.

For comparison with human performance at joint detection of DR and AMD, four independent observers also provided a score for each disease. Two of the graders were certified ophthalmologists with between one and three years of experience (VS, PH) and the other two graders were ophthalmology residents in their last year of residency (YL, AD).

The gradings from the RS and the independent observers were then adjusted for the adaptation of the detection of both diseases into two separate binary classifications. In the case of DR: non-referable DR (stage 0 or 1) and referable DR (stage 2, 3, or 4); for AMD: non-referable AMD (stage 1 or 2) and referable AMD (stage 3 or 4). Cases without both referable DR and referable AMD are referred to as controls from now on. Note that this implies non-referability for both DR and AMD, but other eye diseases might be present.

The reference standard for Messidor was made publicly available when the dataset was originally published, with the subsequent correction of the published errata until the realization of this study in 2018.[46] Medical experts provided the retinopathy grade for each image, consisting of four distinct categories, from 0 to 3, ranging from normal to increasing severity of DR. In order to translate this RS into referable/non-referable classification, images assigned with DR stage 0 or 1 were considered non-referable cases; those with DR stage 2 or 3, referable. For human performance comparison, manual annotations were performed by two independent graders, a general ophthalmologist and a retinal specialist, with 4 and 20 years of DR screening experience, respectively, following the same protocol as the RS.[51]

The reference standard for the AREDS dataset corresponds to the publicly-available grading in AREDS dbGaP, which is based on the AREDS severity scale for AMD described previously.[48,50] These scores were assigned to the images by experts at US grading centers, being consistent with the original AREDS AMD categorization without considering visual acuity.[52] This RS was then adapted following the mentioned procedure into referable and non-referable cases for performing binary classification.

**Table 1** summarizes distribution of disease severity for DR and AMD in the validation datasets regarding the corresponding reference standard.

## Automated grading approach

The DL system under validation uses convolutional neural networks (CNN) for the classification task of grading.[53-57] CNNs are organized in multiple layers with artificial neurons, which learn representations of the input data at increasing levels of abstraction.[36] Convolution operations act as feature detectors with adjustable parameters called weights. During training, the network is presented with a large set of annotated images. For each image, an output class label is produced in a forward pass through the network and a loss function is computed to measure the error between the output and the actual label. With the aim of reducing the error, the weights are adjusted by means of backpropagation.[32] This process is repeated with several passes over the training data until the loss converges.

RetCAD v.1.3.0 consists of two ensembles of state-of-the-art CNN architectures. Both ensembles, correspondingly, allow for the detection of referable DR and AMD. Firstly, each input image goes through a preprocessing stage, followed by an assessment of image quality. Then, joint image-level detection of DR and AMD is applied. Each ensemble provides one score between 0 and 100 which is monotonically related to the likelihood of presence of referable DR and AMD, respectively. The final score for each disease is obtained by averaging the scores generated by the networks in each ensemble.

None of the images included in the datasets used in this validation study were used for training the system.

## Evaluation design

To evaluate the performance of the system at automated joint detection of referable DR and AMD, we performed several validation experiments on the DR-AMD dataset. For detection of referable DR, binary classification was assessed between DR cases and the joint set of controls and AMD cases (DR vs. AMD + controls). A second binary classification for DR was performed between referable and only control cases, in order to analyze the influence of joint AMD cases in the performance of the system (DR vs. controls). The same procedure was applied for detection of referable AMD, assessing first a binary classification between AMD cases and the joint set of controls and DR cases (AMD vs. DR + controls), and a subsequent binary classification between AMD and only control cases (AMD vs. controls).

Regarding validation of individual detection of referable DR and AMD, binary classification was performed between referable and non-referable DR cases in Messidor, and between referable and non-referable AMD cases in the AREDS dataset.

The performance metrics used for validation were sensitivity (SE) and specificity (SP), defined as the proportions of cases considered referable and non-referable, respectively, by both the system and the reference standard. The tradeoff between both metrics was furtherly observed by means of receiver operating characteristic (ROC) analysis. The optimal operating point of the system was considered to be the best tradeoff between SE and SP, i.e., the point closest to the upper left corner of the graph. For an overall interpretation of the system's ability to discriminate between referable and non-referable cases, the area under the ROC curve (AUC) was computed. Human performance was also evaluated by computing sensitivity and specificity from the gradings of each observer and then included in the corresponding ROC curve as operating points.

Data bootstrapping was used to assess statistical significance of the obtained evaluation metrics.[58] Samples were bootstrapped 1000 times to generate a distribution of each evaluation metric, obtaining the 2.5 and 97.5 percentiles as 95% confidence intervals (CI).

Additionally, in the validation datasets where gradings by independent observers were available, i.e., DR-AMD dataset and Messidor, intergrader variability was measured by means of the quadratic Cohen's weighted kappa coefficient (κ), between gradings per disease stage and the corresponding reference standard.[59]

## Results

For the 600 images in the DR-AMD dataset, the ROC analysis corresponding to DR vs. AMD + controls is shown in **Figure 1A**. The optimal operating point of RetCAD v.1.3.0 corresponds to SE of 90.1% (95% CI, 84.2%-96.6%) and SP of 90.6% (95% CI, 85.9%-97.0%), with AUC of 95.1% (95% CI, 90.8%-98.2%). Average observer SE and SP were 61.5% and 97.8%, respectively. **Figure 1B** shows the ROC curve and optimal operating point by the DL system regarding AMD vs. DR + controls. SE was 91.8% (95% CI, 84.4%-97.6%), SP was 87.5% (95% CI, 83.5%-97.9%), and AUC was 94.9% (95% CI, 90.9%-97.9%). Average observer SE and SP were 76.5% and 96.1%, respectively. **Table 2** summarizes diagnostic performance of the system and the human observers for both validation experiments.

Regarding validation of DR vs. controls on the DR-AMD dataset, AUC was 95.6% (95% CI, 91.8%-98.6%), SE was 91.7% (95% CI, 85.3%-98.0%) and SP was 90.9% (95% CI, 86.7%-96.7%). As for AMD vs. controls, AUC was 95.2% (95% CI, 91.0%-98.1%), SE was 88.6% (95% CI, 83.8%-100.0%) and SP was 92.1% (95% CI, 84.3%-95.2%). The corresponding ROC analysis and distribution of both classification results of RetCAD v.1.3.0 and the observers can be found in **Supplementary Figure S2** and **Supplementary Table S1**.

Intergrader disagreement in the DR-AMD dataset is shown in **Figure 2**, which includes interrater heatmaps with quadratic-weighted κ scores among the four observers and the reference standard, for DR and AMD.

Regarding the performance validation of the system and external observers at detection of referable DR in Messidor, the obtained results can be found in **Figure 3A**. The AUC was 97.5% (95% CI, 96.3%-98.5%), SE was 92.0% (95% CI, 89.3%-97.2%) and SP was 92.1% (95% CI, 88.6%-95.2%). Diagnostic performance by the system and the two observers is summarized in **Supplementary Table S2**, while **Supplementary Figure S3** shows the intergrader discrepancy among observers and the reference standard.

The results of the ROC analysis for automated detection of referable AMD in the AREDS dataset are shown in **Figure 3B**. For the 133821 images, the DL system reached 85.8% (95% CI, 84.6%-86.2%) for SE and 86.0% (95% CI, 85.7%-87.4%) for SP. AUC was 92.7% (95% CI, 92.5%-92.9%). The classification results regarding the reference standard can be found in **Supplementary Table S3**.

# Discussion

In this study we validated the performance for joint detection of referable DR and AMD of a commercially-available, CE-certified DL system, RetCAD v.1.3.0 (Thirona, Nijmegen, The Netherlands) and compared it with independent human observers. The results in the DR-AMD dataset show the system is able to differentiate between the two diseases, which is one of the main aspects in joint detection. When identifying referable DR, false positive detections can be divided in cases graded as control or referable AMD in the reference standard, being the latter the 17.4% of the cases wrongly classified as referable DR. Regarding false positive cases at detection of referable AMD, 24.2% were graded as referable DR in the reference standard. Furthermore, the performance of the system is not significantly altered when individual disease detection is assessed on the same dataset.

The outcome of the joint validation also demonstrates the DL system performs comparably to human experts. RetCAD v.1.3.0 reaches lower specificity levels than human average, but higher sensitivity for both DR and AMD. This is particularly important at automated screening settings, where fewer referable cases must be missed when the system is used for either initial assessment or grading support.

Regarding intergrader variability, greater disagreement was observed for AMD, which might show the necessity of establishing AMD screening protocols as the ones already used for DR. For DR, we observed relatively low sensitivity scores for the observers regarding the reference standard, since many of the cases classified as stage 2 in the reference were graded by observers as stage 1. However, interobserver scores are still relatively high. This indicates graded stages are close, but intermediate DR levels become problematic for referable/non-referable classification.

DL-based automated joint detection was also assessed by Ting et al.,[45] reaching lower AUC values at detection of referable DR and AMD, although larger datasets were used for validation and detection of glaucoma was also evaluated. However, fewer external observers were included and different validation sets were used for identification of DR and AMD, which leaves the influence of each disease at joint screening unclear.

Validation of individual detection of referable DR in Messidor shows exceptional performance by RetCAD v.1.3.0, also comparable to human experts. Intermediate DR stages are generally more difficult to identify (97.6% of false negatives belong to stage 2 and 92.7% of false positives belong to stage 1), as noted previously with the human observers in the DR-AMD dataset. Nevertheless, detection errors are kept remarkably low.

Previous DL approaches for DR detection have been reported in Messidor-2[60] with optimal performances. Since there is no publicly-available image-based reference standard for this extension of Messidor, we reported on the original set to allow for further comparison. Gulshan et al.[38] used their own reference standard for Messidor-2, whereas patient-based reference standard was made available and applied by Abramoff et al.[34,40,61] We used this RS for additional validation in Messidor-2 (see **Supplementary Results Appendix**, **Supplementary Figure S4** and **Supplementary Table S4**).

The results of individual detection of referable AMD in the AREDS dataset show that, as with DR, misclassifications shift towards intermediate stages (86.0% of false positives belong to cases graded as AMD stage 2 in the reference standard, whereas 67.1% of false negatives belong to stage 3). RetCAD v.1.3.0 performs at a good level, considering the images in this set are digitized analog photographs. Burlina et al.[42] also reported on DL-based referable AMD detection in the whole AREDS dataset, using the set also for training, which might explain better performance.

Limitations and future work

Although the output score of the validated DL system for DR and AMD is related to the presence of each disease, there is no clear cutoff for disease staging, which could be especially beneficial for easier identification of intermediate stages, since they tend to be more ambiguous to diagnose.

Integrating other imaging modalities such as optical coherence tomography could provide valuable information for diagnosis. However, due to cost-effectiveness and easier adaptation in telemedicine,[62] CF imaging facilitates screening of eye diseases, especially in developing countries.

This validation shows the capacity of a commercially-available DL system to assess joint detection of DR and AMD. However, future integration of automated detection of other eye diseases that might co-exist, such as glaucoma and cataracts, might increase usability and support at screening settings.

With respect to this study, the human observers were professional ophthalmologists or ophthalmologists in training, who are used to clinical working settings and tasks, where the prevalence of disease and manual grading tasks differ from those of real screening settings. Besides, for evaluation of joint detection performance, one DR-AMD set of 600 images from 288 patients was used. In this dataset, patients might contain images from different visits, and in some cases, several images from the same visit. Future studies on automated joint screening would benefit from more and larger validation datasets, with more subjects and increased inter-subject variability, which would allow to analyse the effect of higher patient and imaging diversity on the performance of automated approaches. Additionally, these datasets would be even more beneficial by including graded cases with different severity levels for DR, AMD and additional eye diseases.

In conclusion, this validation study shows the capability of a commercially-available, CE-certified DL system to assess simultaneous detection of DR and AMD with performance comparable to human experts. This demonstrates that an automated solution for joint detection would be beneficial at screening settings, since eye diseases can co-exist and examination protocols rely on the same principles and actions, while reducing subjectivity due to interobserver disagreement. This also shows that DL systems can facilitate access to screening of eye diseases, both in high- and low-resource areas, and become a quick and reliable support for ophthalmological experts.

**Table 1. Disease severity distribution for DR and AMD in the validation datasets.**

| | | Disease stage | DR-AMD*,† | | Messidor‡ | | AREDS† | |
|---|---|---|---|---|---|---|---|---|
| **DR** | **NR** | 0 | 489 (81.5) | 487 (81.2) | 700 (58.3) | 547 (45.6) | - | - |
| | | 1 | | 2 (0.3) | | 153 (12.7) | | - |
| | **R** | 2 | 111 (18.5) | 82 (13.6) | 500 (41.7) | 247 (20.6) | - | - |
| | | 3 | | 4 (0.7) | | 253 (21.1) | | - |
| | | 4 | | 25 (4.2) | | - | | - |
| **AMD** | **NR** | 1 | 527 (87.8) | 483 (80.5) | - | - | 74401 (55.6) | 41409 (30.9) |
| | | 2 | | 44 (7.3) | - | - | | 32992 (24.7) |
| | **R** | 3 | 73 (12.2) | 54 (9.0) | - | - | 59420 (44.4) | 41495 (31.0) |
| | | 4 | | 19 (4.2) | - | - | | 17925 (13.4) |
| **Total images with available grading, No. (%)** | | | 600 (100) | | 1200 (100) | | 133821 (100) | |

\* Reference standard for DR grading is ICDR (stages from 0 to 4)

† Reference standard for AMD grading is AREDS (stages from 1 to 4).

‡ Reference standard for DR grading is Messidor (stages from 0 to 3).

Abbreviations: DR, diabetic retinopathy; AMD, age-related macular degeneration; R, referable; NR, non-referable.

**Figure 1. Receiver operating characteristic curves for joint detection of referable DR (A) and AMD (B) in the DR-AMD dataset (600 images).**

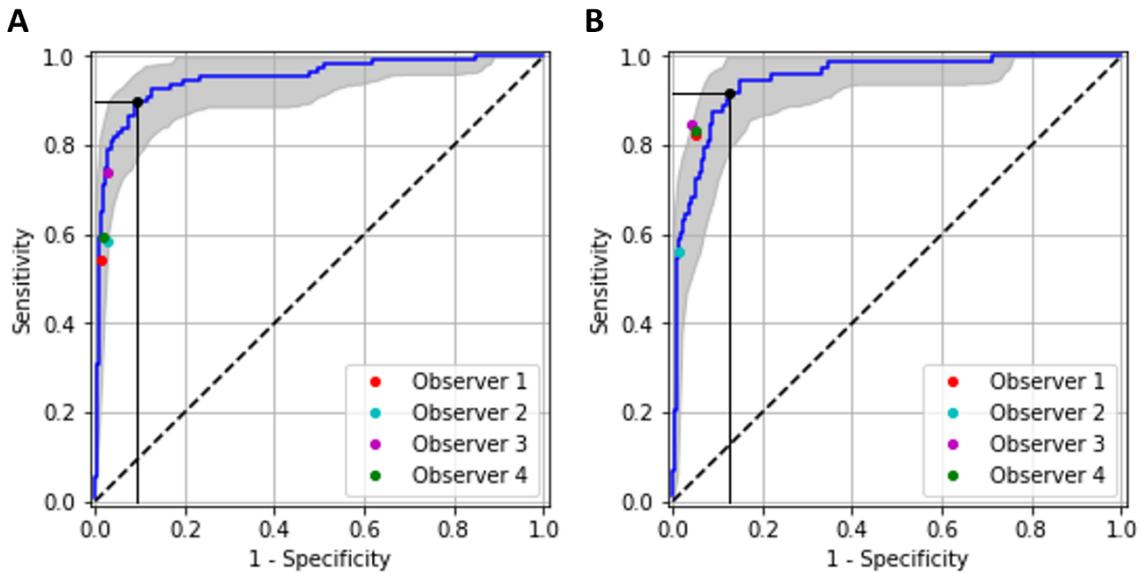

Performance of RetCAD v.1.3.0 corresponds to the blue curves (95% CI within gray area); the colored circles, to the human observers. The black circle indicates the SE and SP of RetCAD v.1.3.0 at its optimal operating point. For DR vs. AMD + controls (A), AUC was 95.1% (95% CI, 90.8%-98.2%), SE was 90.1% (95% CI, 85.2%-96.8%) and SP was 90.6% (95% CI, 85.5%-96.7%). For AMD vs. DR + controls (B), AUC was 94.9% (95% CI, 90.9%-97.9%), SE was 91.8% (95% CI, 84.6%-97.8%) and SP was 87.5% (95% CI, 83.5%-93.9%).

Abbreviations: DR, diabetic retinopathy; AMD, age-related macular degeneration; AUC, area under the receiver operating characteristic curve.; SE, sensitivity; SP, specificity; CI, confidence interval.

**Table 2. Diagnostic performance for joint detection of referable DR and AMD of RetCAD v.1.3.0 and observers compared with reference standard in the DR-AMD dataset (600 images).**

|  |  | RetCAD | | Obs. 1 | | Obs. 2 | | Obs. 3 | | Obs. 4 | |
|---|---|---|---|---|---|---|---|---|---|---|---|
| **DR** |  | **R** | **NR** | **R** | **NR** | **R** | **NR** | **R** | **NR** | **R** | **NR** |
| **RS** | **R** | 99 | 12 | 60 | 51 | 65 | 46 | 82 | 29 | 66 | 45 |
|  | **NR (AMD, C)** | 46 (8, 38) | 443 | 7 (1, 6) | 482 | 14 (4, 10) | 475 | 14 (1, 13) | 475 | 9 (1, 8) | 480 |
|  | **SE (%) (95% CI)** | 90.1 (85.2-96.8) | | 54.1 (40.1-67.3) | | 58.6 (46.3-70.9) | | 73.9 (61.4-85.2) | | 59.5 (46.7-72.0) | |
|  | **SP (%) (95% CI)** | 90.6 (85.5-96.7) | | 98.6 (97.0-99.6) | | 97.1 (94.8-99.2) | | 97.1 (94.7-99.2) | | 98.2 (96.2-99.6) | |
| **AMD** |  | **R** | **NR** | **R** | **NR** | **R** | **NR** | **R** | **NR** | **R** | **NR** |
| **RS** | **R** | 66 | 7 | 60 | 13 | 41 | 32 | 62 | 11 | 60 | 13 |
|  | **NR (DR, C)** | 66 (16, 50) | 461 | 26 (8, 18) | 501 | 8 (3, 5) | 519 | 22 (7, 15) | 505 | 26 (5, 21) | 501 |
|  | **SE (%) (95% CI)** | 91.8 (84.6-97.8) | | 82.1 (69.0-93.8) | | 56.2 (39.5-71.8) | | 84.9 (72.5-96.0) | | 82.1 (69.7-93.5) | |
|  | **SP (%) (95% CI)** | 87.5 (83.5-93.9) | | 95.1 (92.2-97.5) | | 98.5 (96.9-99.6) | | 95.8 (93.1-98.1) | | 95.1 (92.2-97.5) | |

Abbreviations: DR, diabetic retinopathy; AMD, age-related macular degeneration; RS, reference standard; Obs., observer; R, referable; NR, non-referable; C, control; SE, sensitivity; SP, specificity; CI, confidence interval.

**Figure 2. Intergrader disagreement in DR (A) and AMD (B) grading stages among independent human observers and reference standard in the DR-AMD dataset (600 images).**

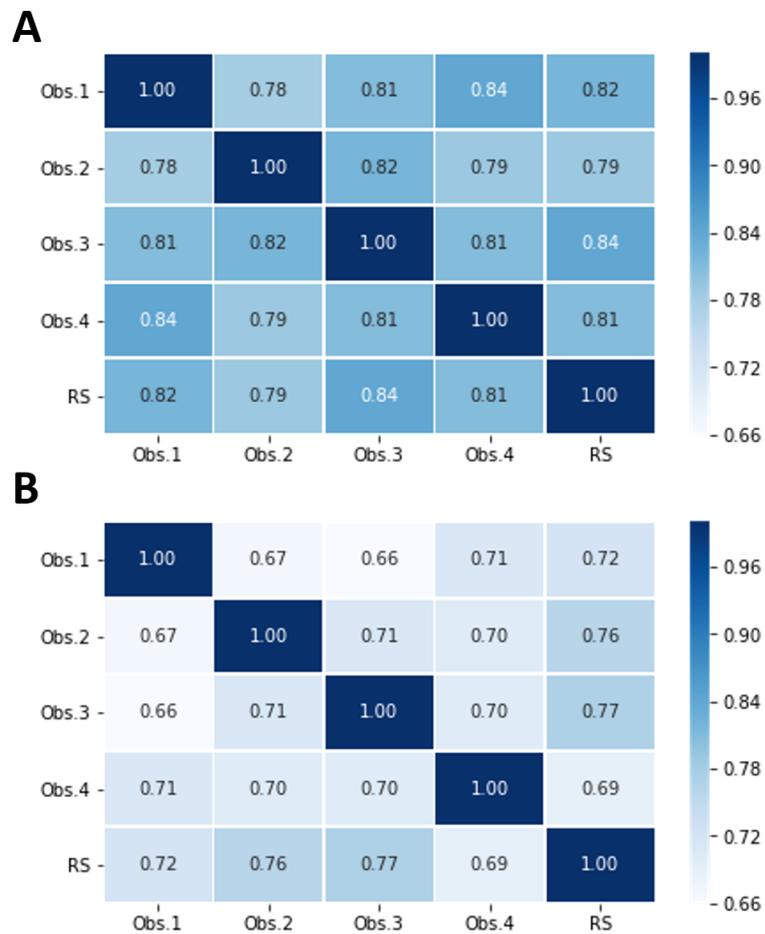

Interrater heatmaps with quadratic Cohen's weighted kappa coefficients comparing disease staging for DR (A) and AMD (B) among the 4 independent human observers and the reference standard in the DR-AMD dataset (600 images with referable DR, AMD and control cases).

Abbreviations: DR, diabetic retinopathy; AMD, age-related macular degeneration; Obs., observer; RS, reference standard.

**Figure 3. Receiver operating characteristic curves for individual detection of referable DR in Messidor (1200 images) (A) and referable AMD in the AREDS dataset (133821 images) (B).**

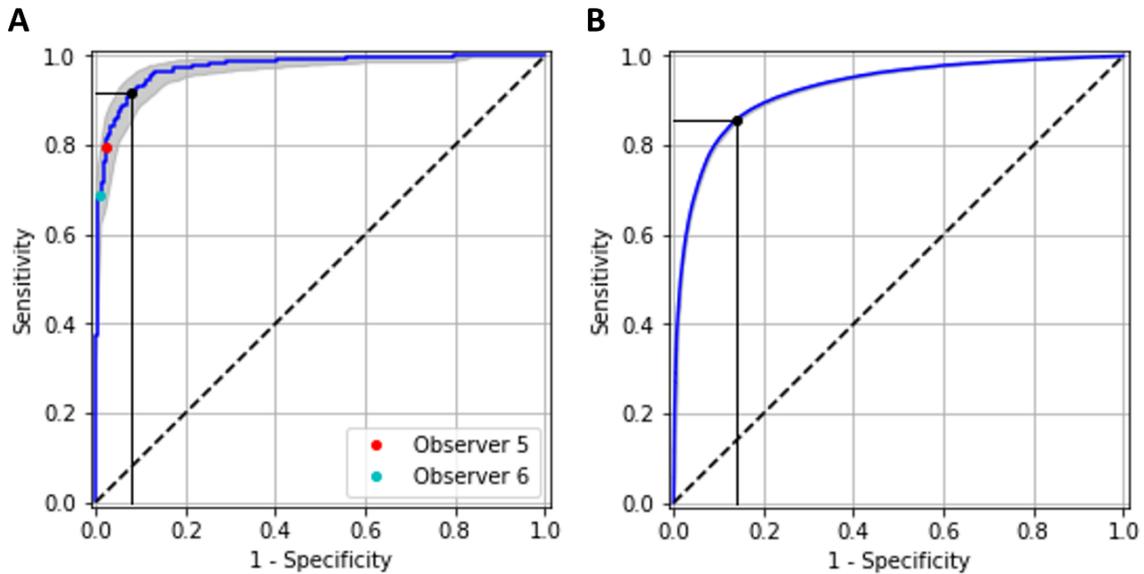

Performance of RetCAD v.1.3.0 corresponds to the blue curves (95% CI within gray area); the colored circles, to the human observers. The black circle indicates the SE and SP of RetCAD v.1.3.0 at its optimal operating point. For individual detection of referable DR (A), AUC was 97.5% (95% CI, 96.3%-98.5%), SE was 92.0% (95% CI, 89.3%-97.2%) and SP was 92.1% (95% CI, 88.6%-95.2%). For individual detection of referable AMD (B), AUC was 92.7% (95% CI, 92.5%-92.9%), SE was 85.8% (95% CI, 84.6%-86.2%) and SP was 86.0% (95% CI, 85.7%-87.4%).

Abbreviations: DR, Diabetic retinopathy; AMD, Age-related macular degeneration; AUC, Area under the receiver operating characteristic curve; SE, sensitivity; SP, specificity; CI, confidence interval.

# Supplementary Online Content

**Evaluation of a deep learning system for the joint automated detection of diabetic retinopathy and age-related macular degeneration**

**Figure S1.** Extraction of the DR-AMD dataset.

**Figure S2.** Receiver operating characteristic curves for individual detection of referable DR and AMD in the DR-AMD dataset.

**Table S1.** Diagnostic performance for individual detection of referable DR and AMD of RetCAD v.1.3.0 and observers compared with reference standard in the DR-AMD dataset.

**Table S2.** Diagnostic performance for individual detection of referable DR of RetCAD v.1.3.0 and observers compared with reference standard in Messidor (1200 images).

**Figure S3.** Interrater disagreement in DR grading stages among independent human observers and reference standard in Messidor (1200 images).

**Table S3.** Diagnostic performance for individual detection of referable AMD of RetCAD v.1.3.0 compared with reference standard in the AREDS dataset (133821 images).

**Results Appendix.** Validation of individual detection of referable DR in Messidor-2.

**Figure S4.** Receiver operating characteristic curve for individual detection of referable DR in Messidor-2 (874 subjects).

**Table S4.** Diagnostic performance for individual detection of referable DR of RetCAD v.1.3.0 compared with reference standard in Messidor-2 (874 subjects).

**Figure S1.** Extraction of the DR-AMD dataset.

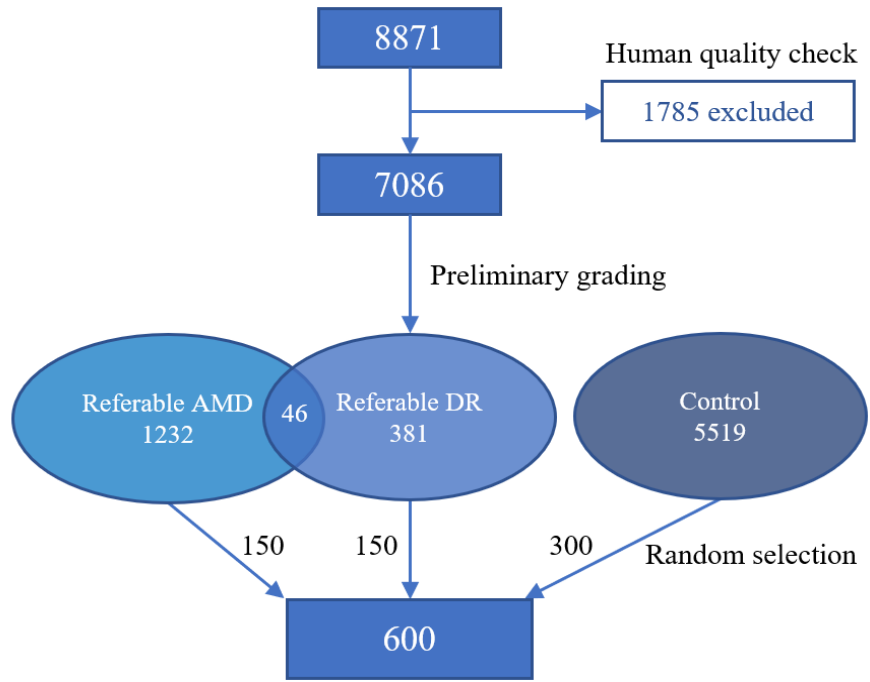

The DR-AMD dataset was extracted from a set of 8871 images from more than 2000 patients collected in three different European medical centers (Sweden, Denmark, Spain) during routine clinical practice. Images went through a human quality check, regarding contrast, clarity and focus, where 1785 images were excluded. The remaining 7086 images went through a preliminary grading, performed by a person with over six years of experience reading CF images. Images were classified as referable AMD (1232 images), referable DR (381 images) or control (5519 images), which indicates non-referability for both DR and AMD, although other diseases might be present; 46 images were graded as having both referable AMD and DR present. Lastly, a random selection of 600 images was performed, containing 150 referable AMD cases, 150 referable DR cases, and 300 controls, in order to ensure an enriched set. These images belong to 288 different patients, with an average of 2.11 images and 1.18 visits per patient. The 600 images define the DR-AMD set used for validation of joint detection of DR and AMD.

Abbreviations: CF, Color fundus; DR, Diabetic retinopathy; AMD, Age-related macular degeneration.

**Figure S2.** Receiver operating characteristic curves for individual detection of referable DR and AMD in the DR-AMD dataset**.**

A. DR vs. controls (527 images): AUC, 95.6% ; 95% CI, 91.8%-98.6%

B. AMD vs. controls (489 images): AUC, 95.2%; 95% CI, 91.0%-98.1%

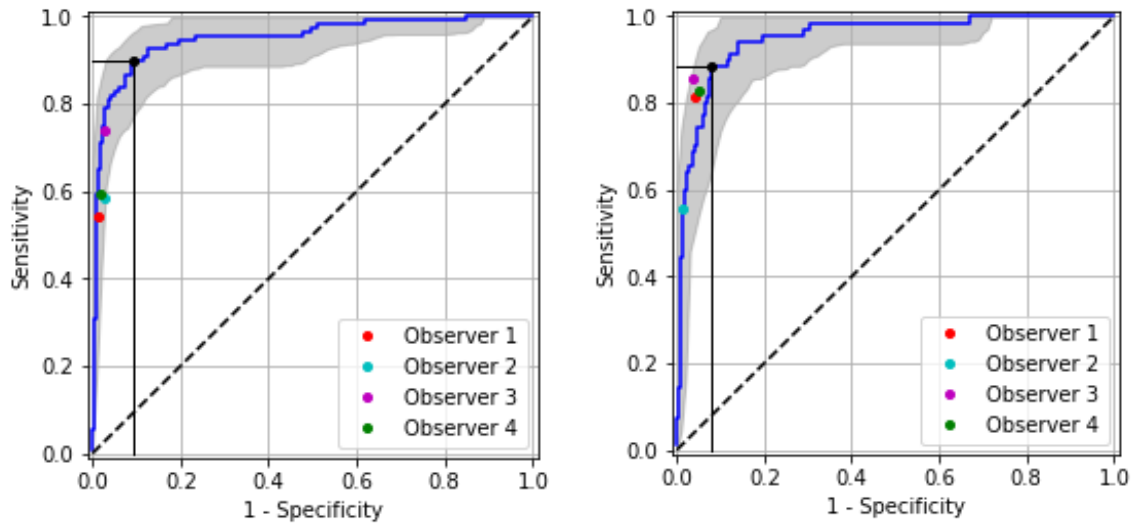

Performance of RetCAD v.1.3.0 (blue curves; 95% CI within gray area) and human observers (colored circles) in the DR-AMD dataset for individual detection of referable DR (527 images with referable DR and control cases) (A) and individual detection of referable AMD (489 images with referable AMD and control cases) (B). The black circle indicates the SE and SP of RetCAD v.1.3.0 at its optimal operating point. In A, SE was 91.7% (95% CI, 85.3%-98.0%) and SP was 90.9% (95% CI, 86.7%-96.7%). In B, SE was 88.6% (95% CI, 83.8%-100.0%) and SP was 92.1% (95% CI, 84.3%-95.2%).

Abbreviations: DR, Diabetic retinopathy; AMD, Age-related macular degeneration; AUC, Area under the receiver operating characteristic curve; SE, sensitivity; SP, specificity; CI, confidence interval.

**Table S1.** Diagnostic performance for individual detection of referable DR and AMD of RetCAD v.1.3.0 and observers compared with reference standard in the DR-AMD dataset.

|  |  | RetCAD |  | Obs. 1 |  | Obs. 2 |  | Obs. 3 |  | Obs. 4 |  |
|---|---|---|---|---|---|---|---|---|---|---|---|
| **DR (527 images)** |  | **R** | **NR** | **R** | **NR** | **R** | **NR** | **R** | **NR** | **R** | **NR** |
| **RS** | **R** | 98 | 10 | 59 | 49 | 64 | 44 | 81 | 27 | 65 | 43 |
|  | **NR** | 38 | 381 | 6 | 413 | 10 | 409 | 13 | 406 | 8 | 411 |
|  | **SE (%) (95% CI)** | 91.7 (85.3-98.0) | | 54.6 (40.4-67.3) | | 59.3 (45.6-72.1) | | 75.0 (63.2-86.4) | | 60.0 (47.1-72.6) | |
|  | **SP (%) (95% CI)** | 90.9 (86.7-96.7) | | 98.6 (96.7-100.0) | | 97.6 (95.4-99.5) | | 96.9 (94.4-99.0) | | 98.1 (96.1-99.5) | |
| **AMD (489 images)** |  | **R** | **NR** | **R** | **NR** | **R** | **NR** | **R** | **NR** | **R** | **NR** |
| **RS** | **R** | 61 | 9 | 57 | 13 | 39 | 31 | 60 | 10 | 58 | 12 |
|  | **NR** | 33 | 386 | 18 | 401 | 5 | 414 | 15 | 404 | 21 | 398 |
|  | **SE (%) (95% CI)** | 88.6 (83.8-100.0) | | 81.4 (66.7-94.3) | | 55.7 (39.0-72.4) | | 85.7 (73.3-96.8) | | 82.9 (70.0-93.9) | |
|  | **SP (%) (95% CI)** | 92.1 (84.3-95.2) | | 95.7 (92.9-98.1) | | 98.8 (97.1-100.0) | | 96.4 (93.7-98.6) | | 95.0 (91.9-97.6) | |

For individual detection of referable DR: 527 images with referable DR and control cases. For individual detection of referable AMD: 489 images with referable AMD and control cases.

Abbreviations: DR, Diabetic retinopathy; AMD, Age-related macular degeneration; RS, Reference standard; Obs., Observer; R, Referable; NR, Non-referable; C, Control; SE, Sensitivity; SP, Specificity; CI, Confidence interval.

**Table S2.** Diagnostic performance for individual detection of referable DR of RetCAD v.1.3.0 and observers compared with reference standard in Messidor (1200 images).

|     |     | DR stage | RetCAD | | Obs. 5 | | Obs. 6 | |
| --- | --- | --- | --- | --- | --- | --- | --- | --- |
|     |     |     | NR | R | NR | R | NR | R |
| RS  | NR  | 0   | 543 | 4 | 542 | 5 | 545 | 2 |
|     |     | 1   | 102 | 51 | 142 | 11 | 149 | 4 |
|     | R   | 2   | 40 | 207 | 92 | 155 | 121 | 126 |
|     |     | 3   | 1 | 252 | 100 | 243 | 34 | 219 |
|     | **SE (%) (95% CI)** | | 92.0 (89.1-95.9) | | 79.6 (74.8-84.8) | | 69.0 (62.9-74.7) | |
|     | **SP (%) (95% CI)** | | 92.1 (88.7-95.2) | | 97.7 (96.0-99.2) | | 99.1 (97.9-100.0) | |

Abbreviations: RS, Reference Standard; DR, Diabetic retinopathy; R, Referable; NR, Non-referable; Obs., Observer; SE, Sensitivity; SP, Specificity; CI, Confidence interval.

**Figure S3.** Interrater disagreement in DR grading stages among independent human observers and reference standard in Messidor (1200 images).

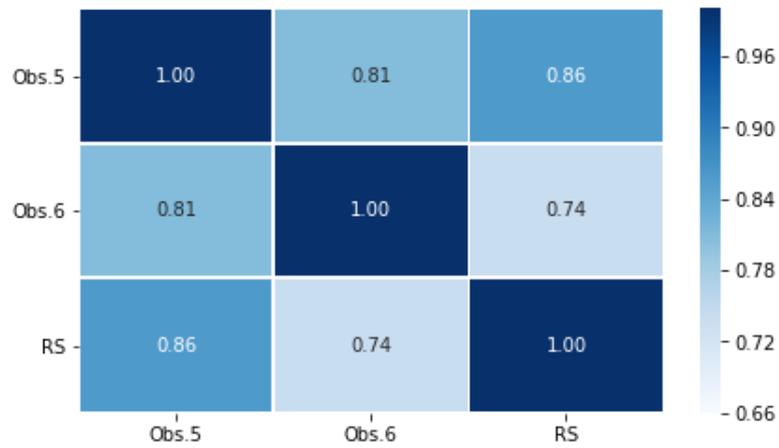

Interrater heatmap with quadratic Cohen's weighted kappa coefficients comparing disease staging for DR among the 2 independent human observers and the reference standard in Messidor (1200 images).

Abbreviations: DR, Diabetic retinopathy; AMD, Age-related macular degeneration; Obs., Observer; RS, Reference Standard.

**Table S3.** Diagnostic performance for individual detection of referable AMD of RetCAD v.1.3.0 compared with reference standard in the AREDS dataset (133821 images).

|     |     |           | RetCAD | |
| --- | --- | --------- | ------ | ------ |
|     |     | AMD stage | NR     | R      |
| RS  | NR  | 1         | 39954  | 1455   |
|     |     | 2         | 24018  | 8974   |
|     | R   | 3         | 7263   | 34232  |
|     |     | 4         | 1146   | 16779  |
|     | SE (%) (95% CI) | | 85.8 (84.6-86.2) | |
|     | SP (%) (95% CI) | | 86.0 (85.7-87.4) | |

Abbreviations: RS, Reference standard; AMD, Age-related macular degeneration; R, Referable; NR, Non-referable; SE, Sensitivity; SP, Specificity; CI, Confidence interval.

**Results Appendix.** Validation of individual detection of referable DR in Messidor-2.

As previous step to generate this dataset, diabetic patients from Brest University Hospital were recruited and new macula-centered paired color fundus (CF) images were obtained without pharmacological dilation, using a Topcon TRC NW6 non-mydriatic fundus camera with a 45-degree field of view. This set of images is known as Messidor-Extension[1] and contains 690 images from 345 examinations, which were combined with the images from the original Messidor set[2] that came in pairs (one image per eye), that is, 1058 images from 529 examinations. The final set is known as Messidor-2[1] and includes in total 1748 images from 874 examinations.

We used as reference standard (RS) the gradings made publicly available by Abramoff et al.[3,4] Three US-board retinal specialists graded each image assigning a level from the International Clinical Diabetic Retinopathy (ICDR) severity scale,[5] obtaining then a consensus between the specialists. The gradings were provided at participant level and regarding referability or non-referability, i.e., a person was classified as a non-referable case if both eyes were classified as level 0 or 1, whereas a participant was considered to have referable DR if one or both eyes belonged to higher stages. In total, 684 subjects (78%) were graded as referable DR and 190 (22%) were graded as non-referable DR.

For the validation of detection of referable DR in Messidor-2 using the mentioned RS, the participant-based score provided by RetCAD v.1.3.0 was computed as the maximum score of both eyes' image-based scores. The DL system achieved an area under the receiver operating characteristic (ROC) curve of 98.0% (95% CI, 96.8%-99.0%), sensitivity was 92.6% (95% CI, 88.4%-97.4%) and specificity was 93.4% (95% CI, 89.9%-97.2%). The corresponding ROC analysis and diagnostic performance of the DL framework can be found in **FigureS3** and **FigureS4** (also available at www.aaojournal.org), respectively.

**Figure S4.** Receiver operating characteristic curve for individual detection of referable DR in Messidor-2 (874 subjects).

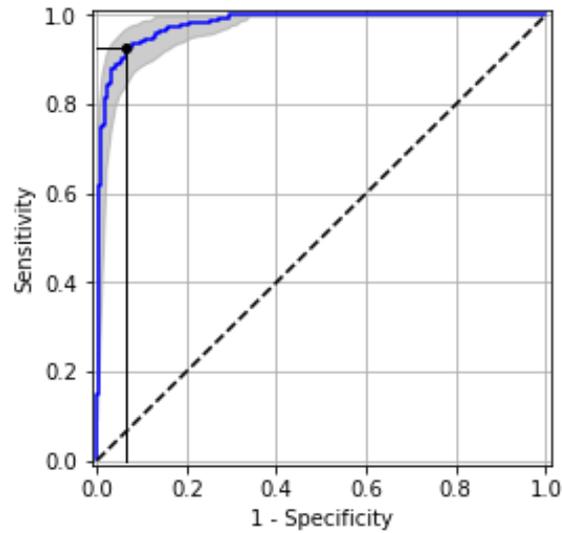

Performance of RetCAD v.1.3.0 (blue curve; 95% CI within gray area) in Messidor-2 (874 subjects) for individual detection of referable DR. The black circle indicates the sensitivity (SE) and specificity (SP) of RetCAD v.1.3.0 at its optimal operating point. SE was 92.6% (95% CI, 88.4%-97.4%) and SP was 93.4% (95% CI, 89.9%-97.2%). AUC was 98.0% % (95% CI, 96.8%-99.0%).

Abbreviations: DR, Diabetic retinopathy; AUC, Area under the receiver operating characteristic curve; SE, Sensitivity; SP, Specificity; CI, Confidence interval.

**Table S4.** Diagnostic performance for individual detection of referable DR of RetCAD v.1.3.0 compared with reference standard in Messidor-2 (874 subjects)**.**

|    |          | RetCAD         |                |
|----|----------|----------------|----------------|
|    | DR label | NR             | R              |
| RS | NR       | 639            | 45             |
|    | R        | 15             | 175            |
|    | SE (%) (95% CI) | 92.6 (88.4-97.4) ||
|    | SP (%) (95% CI) | 93.4 (89.9-97.2) ||

Abbreviations: RS, Reference standard; DR, Diabetic retinopathy; R, Referable; NR, Non-referable; SE, Sensitivity; SP, Specificity; CI, Confidence interval.